# Estimating Classification Uncertainty of Bayesian Decision Tree Technique on Financial Data

Vitaly Schetinin, Jonathan E. Fieldsend, Derek Partridge, Wojtek J. Krzanowski, Richard M. Everson, Trevor C. Bailey and Adolfo Hernandez

School of Engineering, Computer Science and Mathematics, University of Exeter, EX4 4QF, UK

Emails: {v.schetinin, j.e.fieldsend, d.partridge, w.j.krzanowski, r.m.everson, t.c.bailey, a.hernandez}@exeter.ac.uk

**Abstract.** Bayesian averaging over classification models allows the uncertainty of classification outcomes to be evaluated, which is of crucial importance for making reliable decisions in applications such as financial in which risks have to be estimated. The uncertainty of classification is determined by a trade-off between the amount of data available for training, the diversity of a classifier ensemble and the required performance. The interpretability of classification models can also give useful information for experts responsible for making reliable classifications. For this reason Decision Trees (DTs) seem to be attractive classification models. The required diversity of the DT ensemble can be achieved by using the Bayesian model averaging all possible DTs. In practice, the Bayesian approach can be implemented on the base of a Markov Chain Monte Carlo (MCMC) technique of random sampling from the posterior distribution. For sampling large DTs, the MCMC method is extended by Reversible Jump technique which allows inducing DTs under given priors. For the case when the prior information on the DT size is unavailable, the sweeping technique defining the prior implicitly reveals a better performance. Within this Chapter we explore the classification uncertainty of the Bayesian MCMC techniques on some datasets from the StatLog Repository and real financial data. The classification uncertainty is compared within an Uncertainty Envelope technique dealing with the class posterior distribution and a given confidence probability. This technique provides realistic estimates of the classification uncertainty which can be easily interpreted in statistical terms with the aim of risk evaluation.

## 1. Introduction

The uncertainty of Bayesian model averaging used for applications such as financial prediction in which risks should be evaluated is of crucial importance. In general, uncertainty is a triple trade-off between the amount of data available for training, the classifier diversity and the classification accuracy [1 - 4]. The interpretability of classification models can also give useful information to experts responsible for making reliable classifications. For this reason Decision Trees (DTs) seem to be attractive classification models for experts [1 - 7].

The main idea of using DT classification models is to recursively partition data points in an axis-parallel manner. Such models provide natural feature selection and uncover the features which make the important contribution the classification. The resultant DT classification models can be easily interpretable by users.

By definition, DTs consist of splitting and terminal nodes, which are also known as tree leaves. DTs are said to be binary if the splitting nodes ask a specific question and then divide the data points into two disjoint subsets, say the left or the right branch. Fig. 1 depicts an example of the DT consisting of two splitting and three terminal nodes.

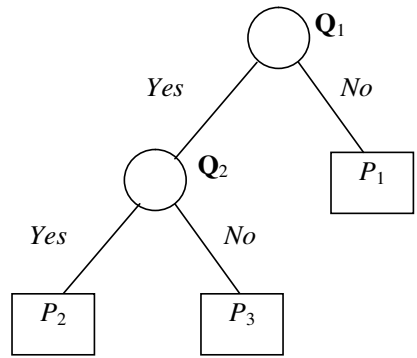

Fig. 1. An example of decision tree consisting of two splitting and terminal nodes depicted by the circles and rectangles. The split nodes ask the questions $Q_1$ and $Q_2$ and an outcome is assigned to one of the terminal nodes with the probabilities $P_1$, $P_2$, and $P_3$.

Note that the number of the data points in each split should not be less than that predefined by a user. The terminal node assigns all data points falling in that node to a class of majority of the training data points resid-ing in this terminal node. Within a Bayesian framework, the class posterior distribution is calculated for each terminal node [4 - 7].

The required diversity of the DTs can be achieved on the base of Bayesian Markov Chain Monte Carlo (MCMC) methodology of sampling from the posterior distribution [4 - 7]. This technique has revealed promising results when applied to some real-world problems. Chipman *et al.* [6] and recently Denison *et al.* [7] have suggested the MCMC techniques in which for sampling from large DTs they used Reversible Jumps (RJ) extension suggested by Green [8]. The RJ MCMC technique making such moves as *birth* and *death* allows the DTs to be induced under the priors given on the shape or size of the DTs. Exploring the posterior distribution, the RJ MCMC should keep the balance between the birth and death moves under which the desired estimate of the posterior can be unbiased [6 - 8].

Within the RJ MCMC technique the proposed moves for which the number of data points falling in one of splitting nodes becomes less than the given number are assigned unavailable. Obviously that the priors given on the DTs are dependent on the class boundaries and noise level in data available for training, and it is intuitively clear that the sharper class boundaries, the larger DTs should be. However in practice the use of such an intuition without a prior knowledge on favourite shape of the DTs can lead to inducing over-complicated DTs and as a result the averaging over such DTs can produce biased class posterior estimates [6, 7]. More over, within the standard RJ MCMC technique suggested for averaging over DTs, the required balance cannot be kept. This may happen because of over-fitting the Bayesian DTs [9]. Another reason is that the RJ MCMC technique averaging over DTs assigns some moves which can not provide a given number of data points allowed being in the splitting nodes unavailable [10].

For the cases when the prior information of the favourite shape of DTs is unavailable, the Bayesian DT technique with a sweeping strategy has revealed a better performance [10]. Within this strategy the prior given on the number of DT nodes is defined implicitly and dependent on the given number of data points allowed being at the DT splits. So the sweeping strategy gives more chances to induce the DTs containing a near optimal number of splitting nodes required to provide the best generalization. At the same time within this technique the number of data points allowed to be in the splitting nodes can be reasonably reduced without increasing the risk of overcomplicating the DTs.

In this Chapter we compare the classification uncertainty of the Bayesian DT techniques with the standard and sweeping RJ MCMC strategies on a synthetic dataset as well on the real financial datasets known as the Aus-tralian and German Credit datasets from the StatLog Repository [11]. In our experiments we also used the Company Liquidity Data recently announced by the German Classification Society for competition in data mining [12]. The classification uncertainty of the Bayesian techniques is evaluated within an Uncertainty Envelope dealing with the class posterior distribution and a given confidence probability suggested in [13]. The Uncertainty Envelope technique by estimating the consistency of DT outputs on the given data produces allows the classification uncertainty to be estimated and interpreted in statistical terms [14]. Using such an evaluation technique in our comparative experiments, we find that the Bayesian DT technique with the sweeping strategy is superior to the standard RJ MCMC technique.

In section 2 we first describe the standard Bayesian RJ MCMC technique and then in section 3 we describe the Bayesian DT technique with the sweeping strategy. In section 4 we briefly describe the Uncertainty Envelope technique used in our experiments for comparison of the classification uncertainty of the two Bayesian DT techniques. The experimental results are presented in section 5, and section 6 concludes the Chapter.

## 2.  The Bayesian Decision Tree Technique

In this section we first present the Bayesian DT technique based on MCMC search methodology and second describe Reversible Jump extension of the MCMC. Finally we discuss the difficulties of sampling large DTs within the RJ MCMC technique.

### 2.1. The Bayesian Averaging over Decision Trees

In general, the predictive distribution we are interested in is written as an integral over parameters $\theta$ of the classification model

$$p(y|\mathbf{x},\mathbf{D}) = \int_{\theta} p(y|\mathbf{x},\theta,\mathbf{D}) p(\theta|\mathbf{D}) d\theta \qquad (1)$$

where $y$ is the predicted class (1, …, $C$), $\mathbf{x} = (x_1, …, x_m)$ is the $m$-dimensional input vector, and $\mathbf{D}$ denotes the given training data.

The integral (1) can be analytically calculated only in simple cases. In practice, part of the integrand in (1), which is the posterior density of $\theta$ conditioned on the data $\mathbf{D}$, $p(\theta|\mathbf{D})$, cannot usually be evaluated. However if values $\theta^{(1)}, \ldots, \theta^{(N)}$ are drawn from the posterior distribution $p(\theta \mid \mathbf{D})$, we can write

$$p(y \mid \mathbf{x}, \mathbf{D}) \approx \sum_{i=1}^{N} p(y \mid \mathbf{x}, \theta^{(i)}, \mathbf{D}) p(\theta^{(i)} \mid \mathbf{D}) = \frac{1}{N} \sum_{i=1}^{N} p(y \mid \mathbf{x}, \theta^{(i)}, \mathbf{D}) \cdot \quad (2)$$

This is the basis of the MCMC technique for approximating integrals [7]. To perform the approximation, we need to generate random samples from $p(\theta \mid \mathbf{D})$ by running a Markov Chain until it has converged to a stationary distribution. After this we can draw samples from this Markov Chain and calculate the predictive posterior density (2).

Let us now define a classification problem presented by data $(\mathbf{x}_i, y_i)$, $i = 1, \ldots, n$, where $n$ is the number of data points and $y_i \in \{1, \ldots, C\}$ is a categorical response. Using DTs for classification, we need to determine the probability $\varphi_{ij}$ with which a datum $\mathbf{x}$ is assigned by terminal node $i = 1, \ldots, k$ to the $j$th class, where $k$ is the number of terminal nodes in the DT. Initially we can assign a $(C - 1)$-dimensional Dirichlet prior for each terminal node so that $p(\varphi_i \mid \theta) = \text{Di}_{C-1}(\varphi_i \mid \alpha)$, where $\varphi_i = (\varphi_{i1}, \ldots, \varphi_{iC})$, $\theta$ is the vector of DT parameters, and $\alpha = (\alpha_1, \ldots, \alpha_C)$ is a prior vector of constants given for all the classes.

The DT parameters are defined as $\theta = (s_i^{\text{pos}}, s_i^{\text{var}}, s_i^{\text{rule}})$, $i = 1, \ldots, k - 1$, where $s_i^{pos}$, $s_i^{var}$ and $s_i^{rule}$ define the *position*, *predictor* and *rule* of each splitting node, respectively. For these parameters the priors can be specified as follows. First we can define a maximal number of splitting nodes, say, $s_{\max} = n - 1$, so $s_i^{pos} \in \{1, \ldots, s_{\max}\}$. Second we draw any of the $m$ predictors from a uniform discrete distribution $U(1, \ldots, m)$ and assign $s_i^{\text{var}} \in \{1, \ldots, m\}$. Finally the candidate value for the splitting variable $x_j = s_i^{\text{var}}$ is drawn from a uniform discrete distribution $U(x_j^{(1)}, \ldots, x_j^{(N)})$, where $N$ is the total number of possible splitting rules for predictor $x_j$, either categorical or continuous.

Such priors allow the exploring of DTs which partition data in as many ways as possible, and therefore we can assume that each DT with the same number of terminal nodes is equally likely [7]. For this case the prior for a complete DT is described as follows:

$$p(\theta) = \left\{ \prod_{i=1}^{k-1} p(s_i^{rule} \mid s_i^{\text{var}}) p(s_i^{\text{var}}) \right\} p(\{s_i^{pos}\}_1^{k-1}). \quad (3)$$

For a case when there is knowledge of the favoured structure of the DT, Chipman *et al.* [6] suggested a generalisation of the above prior – they as-sume the prior probability of further split of the terminal nodes to be dependent on how many splits have already been made above them. For example, for the *i*th terminal node the probability of its splitting is written as

$$p_{split}(i) = \gamma(1+d_i)^{-\delta}, \tag{4}$$

where $d_i$ is the number of splits made above $i$ and $\gamma, \delta \geq 0$ are given constants. The larger $\delta$, the more the prior favours "bushy" trees. For $\delta = 0$ each DT with the same number of terminal nodes appears with the same prior probability.

Having set the priors on the parameters $\varphi$ and $\theta$, we can determine the marginal likelihood for the data given the classification tree. In the general case this likelihood can be written as a multinomial Dirichlet distribution [7]:

$$p(\mathbf{D} \mid \boldsymbol{\theta}) = \left[\frac{\Gamma\{\alpha C\}}{\{\Gamma(\alpha)\}^C}\right]^k \prod_{i=1}^{C} \frac{\prod_{j}^{C} \Gamma(m_{ij} + \alpha_j)}{\Gamma(n_i + \sum_{j=1}^{C} \alpha_j)}, \tag{5}$$

where $n_i$ is the number of data points falling in the *i*th terminal node of which $m_{ij}$ points are of class *j* and $\Gamma$ is a Gamma function.

## 2.2 Reversible Jumps Extension

To allow sampling DT models of variable dimensionality, the MCMC technique exploits the Reversible Jump extension [8]. This extension allows the MCMC technique to sample large DTs induced from real-world data. To implement the RJ MCMC technique Chipman *et al.* [6] and Denison *et al.* [7] have suggested exploring the posterior probability by using the following types of moves.

- *Birth*. Randomly split the data points falling in one of the terminal nodes by a new splitting node with the variable and rule drawn from the corresponding priors.
- *Death*. Randomly pick a splitting node with two terminal nodes and assign it to be one terminal with the united data points.
- *Change-split*. Randomly pick a splitting node and assign it a new splitting variable and rule drawn from the corresponding priors.
- *Change-rule*. Randomly pick a splitting node and assign it a new rule drawn from a given prior.

The first two moves, *birth* and *death*, are reversible and change the dimensionality of **θ** as described in [7]. The remaining moves provide jumps within the current dimensionality of **θ**. Note that the *change-split* move is included to make "large" jumps which potentially increase the chance of sampling from a maximal posterior whilst the *change-rule* move does "local" jumps.

For the birth moves, the proposal ratio *R* is written

$$R = \frac{q(\boldsymbol{\theta}|\boldsymbol{\theta}')p(\boldsymbol{\theta}')}{q(\boldsymbol{\theta}'|\boldsymbol{\theta})p(\boldsymbol{\theta})}, \tag{6}$$

where the $q(\boldsymbol{\theta}|\boldsymbol{\theta}')$ and $q(\boldsymbol{\theta}'|\boldsymbol{\theta})$ are the proposed distributions, **θ**´ and **θ** are (*k* + 1) and *k*-dimensional vectors of DT parameters, respectively, and *p*(**θ**) and *p*(**θ**´) are the probabilities of the DT with parameters **θ** and **θ**´:

$$p(\boldsymbol{\theta}) = \{\prod_{i=1}^{k-1} \frac{1}{N(s_i^{\text{var}})} \frac{1}{m}\} \frac{k}{S_k} \frac{1}{K}, \tag{7}$$

where $N(s_i^{\text{var}})$ is the number of possible values of $s_i^{\text{var}}$ which can be assigned as a new splitting rule, $S_k$ is the number of ways of constructing a DT with *k* terminal nodes, and *K* is the maximal number of terminal nodes, $K = n - 1$.

For binary DTs, as given from graph theory, the number $S_k$ is the *Catalan number*

$$S_k = \frac{1}{k+1}\binom{2k}{k}, \tag{8}$$

and we can see that for $k \geq 25$ this number becomes astronomically large, $S_k \geq (4.8)^{12}$.

The proposal distributions are as follows

$$q(\boldsymbol{\theta}|\boldsymbol{\theta}') = \frac{d_{k+1}}{D_{Q'}}, \tag{9}$$

$$q(\boldsymbol{\theta}'|\boldsymbol{\theta}) = \frac{b_k}{k} \frac{1}{N(s_k^{\text{var}})} \frac{1}{m}, \tag{10}$$

where $D_{Q1} = D_Q + 1$ is the number of splitting nodes whose branches are both terminal nodes.

Then the proposal ratio for a *birth* is given by

$$R = \frac{d_{k+1}}{b_k} \frac{k}{D_{Q1}} \frac{S_k}{S_{k+1}}. \qquad (11)$$

The number $D_{Q1}$ in (11) is dependent on the DT structure and it is clear that $D_{Q1} < k \ \forall \ k = 1, \ldots, K$. Analysing (11), we can also assume $d_{k+1} = b_k$. Then letting the DTs grow, i.e., $k \to K$, and considering $S_{k+1} > S_k$, we can see that the value of $R \to c$, where $c$ is a constant lying between 0 and 1.

Alternatively, for the death moves the proposal ratio is written as

$$R = \frac{b_k}{d_{k-1}} \frac{D_Q}{(k-1)} \frac{S_k}{S_{k-1}}, \qquad (12)$$

and we can see that under the assumptions considered for the birth moves, $R \geq 1$.

## 2.3. The Difficulties of Sampling Decision Trees

The RJ MCMC technique starts drawing samples from a DT consisting of one splitting node whose parameters were randomly assigned within the predefined priors. So we need to run the Markov Chain while it grows and its likelihood is unstable. This phase is said *burn-in* and it should be preset enough long in order to stabilize the Markov Chain. When the Markov Chain will be enough stable, we can start sampling. This phase is said *post burn-in*.

It is important to note that the DTs grow very quickly during the first burn-in samples. This happens because an increase in log likelihood value for the birth moves is much larger than that for the others. For this reason almost every new partition of data is accepted. Once a DT has grown the *change* moves are accepted with a very small probability and, as a result, the MCMC algorithm tends to get stuck at a particular DT structure instead of exploring all possible structures.

The size of DTs can rationally decrease by defining a minimal number of data points, $p_{min}$, allowed to be in the splitting nodes [3 - 5]. If the number of data points in new partitions made after the birth or change moves becomes less than a given number $p_{min}$, such moves are assigned unavailable, and the RJ MCMC algorithm resamples such moves.

However, when the moves are assigned unavailable, this distorts the proposal probabilities $p_b$, $p_d$, and $p_c$ given for the birth, death, and change moves, respectively. The larger the DT, the smaller the number of data points falling in the splitting nodes, and correspondingly the larger is the probability with which moves become unavailable. Resampling the unavailable moves makes the balance between the proposal probabilities biased.

To show that the balance of proposal probabilities can be biased, let us assume an example with probabilities $p_b$, $p_d$, and $p_c$ set equal to 0.2, 0.2, and 0.6, respectively, note that $p_b + p_d + p_c = 1$. Let the DTs be large so that the birth and change moves are assigned unavailable with probabilities $p_{bu}$ and $p_{cu}$ equal to 0.1 and 0.3, respectively. As a result, the birth and change moves are made with probabilities equal to $(p_b - p_{bu})$ and $(p_c - p_{cu})$, respectively.

Let us now emulate 10000 moves with the given proposal probabilities. The resultant probabilities are shown in Fig. 2.

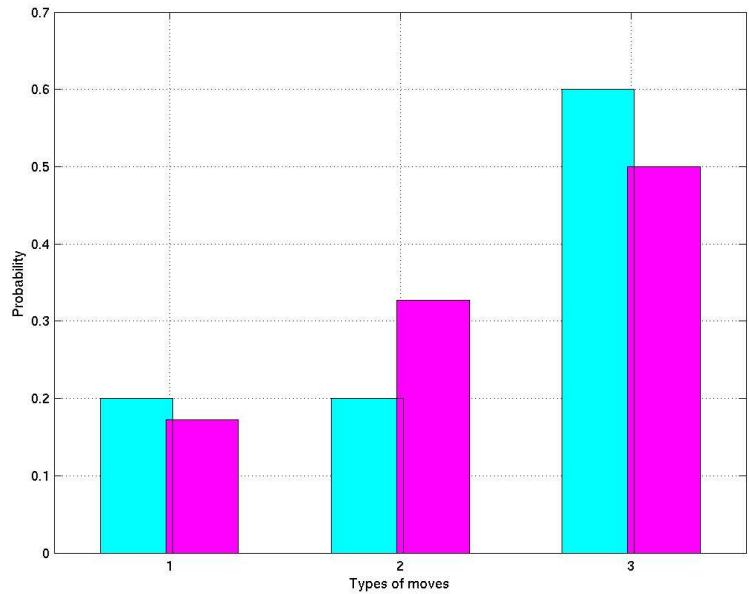

Fig. 2. The standard strategy: The proposal probabilities for the birth, death and change moves presented by the three groups. The left hand bars in each group denote the proposal probabilities. The right hand bars denote the resultant probabilities with which the birth, death, and change moves are made in reality if the birth and change moves were assigned unavailable with probabilities 0.1 and 0.3, respectively.

From the above figure we can see that after resampling the unavailable proposals the probabilities of the birth and death moves become equal ap-proximately 0.17 and 0.32, i.e., the death moves are made with a probability which is significantly larger than a probability originally set equal 0.2.

The disproportion in the balance between the probabilities of birth and death moves is dependent on the size of DTs averaged over samples. Clearly, at the beginning of burn-in phase the disproportion is close to zero, and to the end of the burn-in phase, when the size and form of DTs are stabilized, its value becomes maximal.

Because DTs are hierarchical structures, the changes at the nodes located at the upper levels can significantly change the location of data points at the lower levels. For this reason there is a very small probability of changing and then accepting a DT split located near a root node. Therefore the RJ MCMC algorithm collects the DTs in which the splitting nodes located far from a root node were changed. These nodes typically contain small numbers of data points. Subsequently, the value of log likelihood is not changed much, and such moves are frequently accepted. As a result, the RJ MCMC algorithm cannot explore a full posterior distribution properly.

One way to extend the search space is to restrict DT sizes during a given number of the first burn-in samples as described in [7]. Indeed, under such a restriction, this strategy gives more chances of finding DTs of a smaller size which could be competitive in term of the log likelihood values with the larger DTs. The restricting strategy, however, requires setting up in an *ad hoc* manner the additional parameters such as the size of DTs and the number of the first burn-in samples. Sadly, in practice, it often happens that after the limitation period the DTs grow quickly again and this strategy does not improve the performance.

Alternatively to the above approach based on the explicit limitation of DT size, the search space can be extended by using a restarting strategy as Chipman *et al.* have suggested in [6]. Clearly, both these strategies cannot guarantee that most of DTs will be sampled from a model space region with a maximal posterior. In the next section we describe our approach based on sweeping the DTs.

## 3. The Bayesian Averaging with a Sweeping Strategy

In this section we describe our approach to decreasing the uncertainty of classification outcomes within the Bayesian averaging over DT models. The main idea of this approach is to assign the prior probability of further splitting DT nodes to be dependent on the range of values within which the number of data points will be not less than a given number of points, $p_{min}$.

Such a prior is explicit because at the current partition the range of such values is unknown.

Formally, the probability $P_s(i, j)$ of further splitting at the $i$th partition level and variable $j$ can be written as

$$P_s(i, j) = \frac{x_{\max}^{(i,j)} - x_{\min}^{(i,j)}}{x_{\max}^{(1,j)} - x_{\min}^{(1,j)}}, \quad (13)$$

where $x_{\min}^{(i,j)}$ and $x_{\max}^{(i,j)}$ are the minimal and maximal values of variable $j$ at the $i$th partition level.

Observing Eq. (13), we can see that $x_{\max}^{(i,j)} \leq x_{\max}^{(1,j)}$ and $x_{\min}^{(i,j)} \geq x_{\max}^{(1,j)}$ for all the partition levels $i > 1$. On the other hand there is partition level $k$ at which the number of data points becomes less than a given number $p_{min}$. Therefore, we can conclude that the prior probability of splitting $P_s$ ranges between 0 and 1 for any variable $j$ and the partition levels $i$: $1 \leq i < k$.

From Eq. (13) it follows that for the first level of partition, probability $P_s$ is equal to 1.0 for any variable $j$. Let us now assume that the first partition split the original data set into two non-empty parts. Each of these parts contains less data points than the original data set, and consequently for the ($i = 2$)th partition either $x_{\max}^{(i,j)} < x_{\max}^{(1,j)}$ or $x_{\min}^{(i,j)} > x_{\max}^{(1,j)}$ for new splitting variable $j$. In any case, numerator in (13) decreases, and probability $P_s$ becomes less than 1.0. We can see that each new partition makes values of numerator and consequently probability (13) smaller. So the probability of further splitting nodes is dependent on the level $i$ of partitioning data set.

The above prior favours splitting the terminal nodes which contain a large number of data points. This is clearly a desired property of the RJ MCMC technique because it allows accelerating the convergence of Markov chain. As a result of using prior (13), the RJ MCMC technique of sampling DTs can explore an area of a maximal posterior in more detail.

However, prior (13) is dependent not only on the level of partition but also on the distribution of data points in the partitions. Analyzing the data set at the $i$th partition, we can see that value of probability $P_s$ is dependent on the distribution of these data. For this reason the prior (13) cannot be implemented explicitly without the estimates of the distribution of data points in each partition.

To make the birth and change moves within prior (13), the new splitting values $s_i^{\text{rule,new}}$ for the $i$th node and variable $j$ are assigned as follows. For the birth and change-split moves the new value $s_i^{\text{rule,new}}$ is drawn from a uniform distribution:

$$s_i^{rule,new} \sim U(x_{\min}^{1,j}, x_{\max}^{1,j}). \tag{14}$$

The above prior is "uninformative" and used when no information on preferable values of $s_i^{rule}$ is available. As we can see, the use of a uniform distribution for drawing new rule $s_i^{rule,new}$, proposed at the level $i > 1$, can cause the partitions containing less the data points than $p_{min}$. However, within our technique such proposals can be avoided.

For the change-split moves, drawing $s_i^{rule,new}$ follows after taking new variable $s_i^{var,new}$:

$$s_i^{var,new} \sim U\{S_k\}, \tag{15}$$

where $S_k = \{1, \ldots, m\} \backslash s_i^{var}$ is the set of features excluding variable $s_i^{var}$ currently used at the $i$th node.

For the change-rule moves, the value $s_i^{rule,new}$ is drawn from a Gaussian with a given variance $\sigma_j$:

$$s_i^{rule,new} \sim N(s_i^{rule}, \sigma_j), \tag{16}$$

where $j = s_i^{var}$ is the variable used at the $i$th node.

Because DTs have hierarchical structure, the change moves (especially change-split moves) applied to the first partition levels can heavily modify the shape of the DT, and as a result, its bottom partitions can contain less the data points than $p_{min}$. As mentioned in section 2, within the Bayesian DT techniques [6, 7] such moves are assigned unavailable.

Within our approach after birth or change move there arise three possible cases. In the first case, the number of data points in each new partition is larger than $p_{min}$. The second case is where the number of data points in one new partition is larger than $p_{min}$. The third case is where the number of data points in two or more new partitions is larger than $p_{min}$. These three cases are processed as follows.

For the first case, no further actions are made, and the RJ MCMC algorithm runs as usual.

For the second case, the node containing unacceptable number of data points is removed from the resultant DT. If the move was of birth type, then the RJ MCMC resamples the DT. Otherwise, the algorithm performs the death move.

For the last case, the RG MCMC algorithm resamples the DT.

As we can see, within our approach the terminal node, which after making the birth or change moves contains less than $p_{min}$ data points, is re-moved from the DT. Clearly, removing such unacceptable nodes turns the random search in a direction in which the RJ MCMC algorithm has more chances to find a maximum of the posterior amongst shorter DTs. As in this process the unacceptable nodes are removed, we named such a strategy *sweeping*.

After change move the resultant DT can contain more than one nodes splitting less than $p_{min}$ data points. However this can happen at the beginning of burn-in phase, when the DTs grow, and this unlikely happen, when the DTs have grown.

As an example, Fig. 3 provides the resultant probabilities estimated on 10000 moves for a case when the original probabilities of the birth, death, and change moves were set equal 0.2, 0.2, and 0.6, respectively, as assumed at the example given in section II. The probabilities of the unacceptable birth and change moves were set equal to 0.07 and 0.2. These values are less than those that were set in the previous example because the DTs induced with a sweeping strategy are shorter than those induced with the standard strategy. The shorter DTs, the more data points fall at their splitting nodes, and less the probabilities $p_{bu}$ and $p_{cu}$ are. In addition, 1/10th of the unacceptable change moves was set assigned to the third option, mentioned above, for which two or more new partitions contain less than $p_{min}$ data points.

From Fig. 3 we can see that after resampling the unacceptable birth moves and reassigning the unacceptable change moves, the resultant probabilities of the birth and death moves become equal approximately 0.17 and 0.3, i.e., the values of these probabilities are very similar to those that shown in Fig. 2.

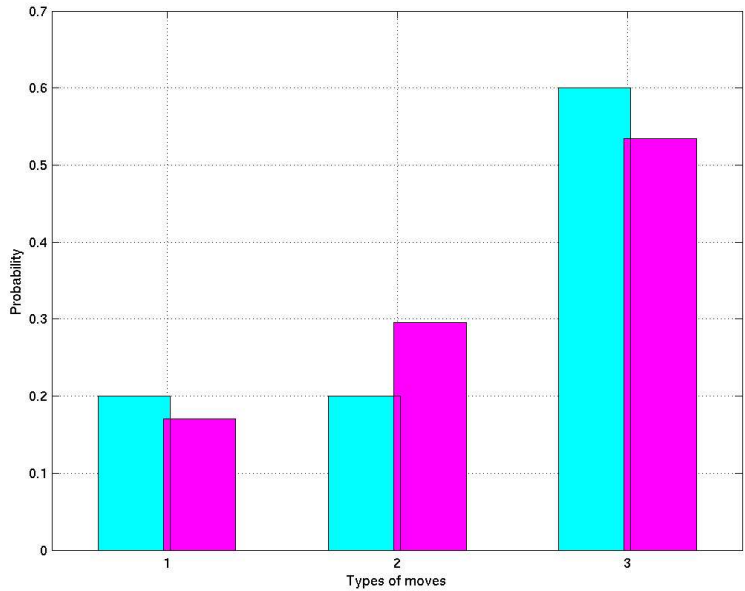

Fig. 3. The shrinking strategy: The proposal probabilities for the birth, death and change moves presented by the three groups. The left hand bars in each group denote the proposal probabilities. The right hand bars denote the resultant probabilities with which the birth, death, and change moves are made in reality if the birth and change moves were assigned unavailable with probabilities 0.07 and 0.2, respectively.

Next we describe the Uncertainty Envelope technique suggested to estimate the classification uncertainty of multiple classifier systems the details of which are described in [13]. This technique allows us to compare the performance of the Bayesian strategies of averaging over the DTs in terms of classification uncertainty.

## 4. The Uncertainty Envelope Technique

In general, the Bayesian DT strategies described in sections 2 and 3 allow sampling the DTs induced from data independently. In such a case, we can naturally assume that the inconsistency of the classifiers on a given datum $\mathbf{x}$ is proportional to the uncertainty of the DT ensemble. Let the value of class posterior probability $P(c_j|\mathbf{x})$ calculated for class $c_j$ be an average over the class posterior probability $P(c_j|K_i, \mathbf{x})$ given on classifier $K_i$:

$$P(c_j \mid \mathbf{x}) = \frac{1}{N} \sum_{i=1}^{N} P(c_j \mid K_i, \mathbf{x}), \quad (17)$$

where $N$ is the number of classifiers in the ensemble.

As classifiers $K_1, \ldots, K_N$ are independent each other and their values $P(c_j|K_i, \mathbf{x})$ range between 0 and 1, the probability $P(c_j|\mathbf{x})$ can be approximated as follows

$$P(c_j \mid \mathbf{x}) \approx \frac{1}{N} \sum_{i=1}^{N} I(y_i, t_i \mid \mathbf{x}), \quad (18)$$

where $I(y_i, t_i)$ is the indicator function assigned to be 1 if the output $y_i$ of the $i$th classifier corresponds to target $t_i$, and 0 if it does not.

The larger number of classifiers, $N$, the smaller is error of the approximation (17). For example, when $N = 500$, the approximation error is equal to 1%, and when $N = 5000$, it becomes equal to 0.4%.

It is important to note that the right side of Eq. (18) can be considered as a *consistency* of the outcomes of DT ensemble. Clearly, values of the consistency, $\gamma = \frac{1}{N} \sum_{i=1}^{N} I(y_i, t_i \mid \mathbf{x})$, lie between 1/C and 1.

Analyzing Eq. (18), we can see that if all the classifiers are degenerate, i.e., $P(c_j|K_i, \mathbf{x}) \in \{0, 1\}$, then the values of $P(c_j|\mathbf{x})$ and $\gamma$ become equal. The outputs of classifiers can be equal to 0 or 1, for example, when the data points of two classes do not overlap. In other cases, the class posterior probabilities of classifiers range between 0 and 1, and the $P(c_j|\mathbf{x}) \approx \gamma$. So we can conclude that the classification confidence of an outcome is characterized by the consistency of the DT ensemble calculated on a given datum. Clearly, the values of $\gamma$ are dependent on how representative the training data are, what classification scheme is used, how well the classifiers were trained within a classification scheme, how close the datum $\mathbf{x}$ is to the class boundaries, how the data are corrupted by noise, and so on.

Let us now consider a simple example of a DT ensemble consisting of $N = 1000$ classifiers in which 2 classifiers give a conflicting classification on a given datum $\mathbf{x}$ to the other 998. Then consistency $\gamma = 1 - 2/1000 = 0.998$, and we can conclude that the DT ensemble was trained well and/or the data point $\mathbf{x}$ lies far from the class boundaries. It is clear that for new datum appearing in some neighbourhood of the $\mathbf{x}$, the classification uncertainty as the probability of misclassification is expected to be $1 - \gamma = 1 - 0.998 = 0.002$. This inference is truthful for the neighbourhood within which the prior probabilities of classes remain the same. When the value of $\gamma$ is close to $\gamma_{min} = 1/C$, the classification uncertainty is highest and a datum **x** can be misclassified with a probability $1 - \gamma = 1 - 1/C$.

From the above consideration, we can assume that there is some value of consistency $\gamma_0$ for which the classification outcome is confident, that is the probability with which a given datum **x** could be misclassified is small enough to be acceptable. Given such a value, we can now specify the uncertainty of classification outcomes in statistical terms. The classification outcome is said to be *confident and correct*, when the probability of misclassification is acceptably small and $\gamma \geq \gamma_0$.

Additionally to the confident and correct output, we can specify a *confident but incorrect* output referring to a case when almost all the classifiers assign a datum **x** to a wrong class whilst $\gamma \geq \gamma_0$. Such outcomes tell us that the majority of the classifiers fail to classify a datum **x** correctly. The confident but incorrect outcomes can happen for different reasons, for example, the datum **x** could be mislabelled or corrupted, or the classifiers within a selected scheme cannot distinguish the data **x** properly.

The remaining cases for which $\gamma < \gamma_0$ are regarded as *uncertain classifications*. In such cases the classification outcomes cannot be accepted with a given confidence probability $\gamma_0$ and the DT ensemble labels them as uncertain.

Fig. 4 gives a graphical illustration for a simple two-class problem formed by two Gaussian $N(0, 1)$ and $N(2, 0.75)$ for variable *x*. As the class probability distributions are given, an optimal decision boundary can be easily calculated in this case. For a given confident consistency $\gamma_0$, the integration over the class posterior distribution gives boundaries B1 and B2 within which the outcomes of the DT ensemble are assigned within the Uncertainty Envelope technique to be confident and correct (CC), confident but incorrect (CI) or uncertain (U). If a decision boundary within a selected classification scheme is not optimal, the classification error becomes higher than a minimal Bayes error. So, for the Bayesian classifier and a given consistency $\gamma_0$, the probabilities of CI and U outcomes on the given data are minimal as depicted in Fig. 4.

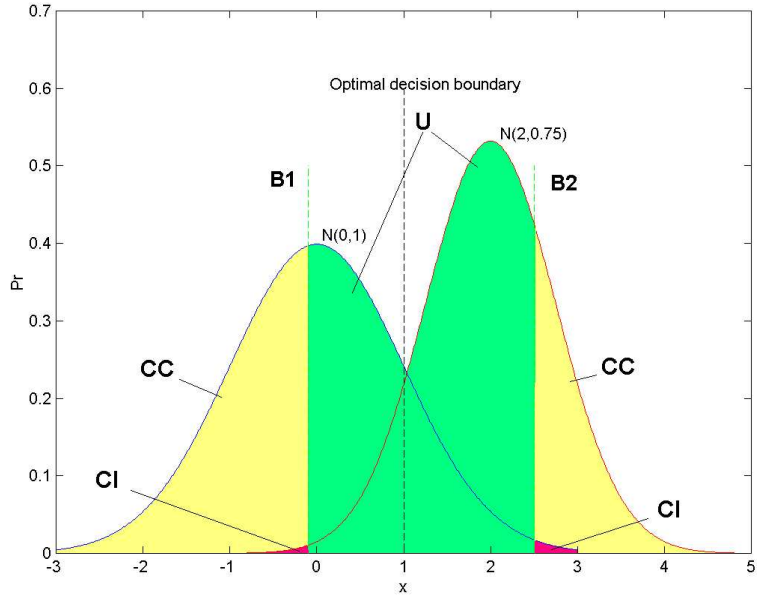

Fig. 4: Uncertainty Envelope characteristics for an example of two-class problem

The above three characteristics, the confident and correct, confident but incorrect, and uncertain outcomes, seem to provide a practical way of evaluating different types of DT ensembles on the same data sets. Comparing the ratios of the data points assigned to be one of these three types of classification outcomes, we can quantitatively evaluate the classification uncertainty of the DT ensembles. Depending on the costs of types of misclassifications in real-world applications, the value of the confidence consistency $\gamma_0$ should be given, say, equal to 0.99.

Next we describe the experimental results obtained with the shrinking strategy of Bayesian averaging over DTs. These results are then compared with those that have been obtained with the standard Bayesian DT technique described in [7].

## 5. Experiments and Results

This section describes the experimental results on the comparison of the Bayesian DT techniques with the standard and sweeping strategies described in the above sections. The experiments were conducted first on a synthetic dataset, and then on the real financial datasets, the Australian and German Credit Datasets available at the StatLog Repository [11] as well as the Company Liquidity Data recently presented by the German Classification Society at [12]. The performance of the Bayesian techniques is evaluated within the Uncertainty Envelope technique described in section 4.

## 5.1. The Characteristics of Datasets and Parameters of MCMC Sampling

The synthetic data are related to an exclusive OR problem (XOR3) with the output $y = \text{sign}(x_1 x_2)$ and three input variables $x_1, x_2 \sim U(-0.5, 0.5)$ and $x_3 \sim N(0, 0.2)$ which is a Gaussian noise. Table 1 lists the total number of input variables, $m$, including the number of the nominal variables, $m_0$, the number of examples, $n$, and the proportion of examples of class 1, $r$. All the four datasets present the two-class problems.

Table 1. The characteristics if the data sets

| # | Data | $m$ | $m_0$ | $n$ | $r,\%$ |
|---|------|-----|-------|-----|--------|
| 1 | XOR3 | 3 | 0 | 1000 | 50.0 |
| 2 | Australian Credit | 14 | 13 | 690 | 55.5 |
| 3 | German Credit | 20 | 20 | 1000 | 70.0 |
| 4 | Company Liquidity | 26 | 15 | 20000 | 88.8 |

Variables with the enumerated number of values were assigned nominal. All the above data do not contain missing values. However the Company Liquidity Data contain many values marked by 9999999 that we interpreted as unimportant under the certain circumstances. The fraction of such values is large and equal 24%.

For all the above domain problems, no prior information on the preferable DT shape and size was available. The pruning factor, or the minimal number of data point allowed being in the splits, $p_{min}$ was given equal between 3 and 50 in the dependence on the size of the data. The proposal probabilities for the death, birth, change-split and change-rules are set to be 0.1, 0.1, 0.2, and 0.6, respectively. The numbers of burn-in and post burn-in samples were also dependent on the problems. Meanwhile, the sampling rate for all the domain problems was set equal to 7. Note all the parameters of MCMC sampling were set the same for both Bayesian techniques.

The performance of the Bayesian MCMC techniques was evaluated within the Uncertainty Envelope techniques within 5 fold cross-validation and $2\sigma$ intervals. The average size of the induced DTs is an important characteristic of the Bayesian techniques and it was also evaluated in our experiments.

### 5.2. Experimental Results

#### 5.2.1. Performance on XOR3 Data

Both Bayesian DT techniques with the standard (DBT1) and the sweeping (BDT2) strategies perform quite well on the XOR3 data, recognizing 99.7% and 100.0% of the test examples, respectively. The acceptance rate was 0.49 for the BDT1 and 0.12 for BDT2 strategies. The average number of DT nodes was 11.3 and 3.4 for these strategies, respectively, see Table 2. Both the BDT1 and the BDT2 strategies ran with the value $p_{min}$ = 5. The numbers of burn-in and post burn-in samples were set equal to 50000 and 10000, respectively. The proposal variance was set equal 1.0.

Table 2: Comparison between BDT1 and BDT2 on the XOR3 Data

| Strategy | Number of DT nodes | Perform, % | Sure correct, % | Uncertain, % | Sure incorrect, % |
|---|---|---|---|---|---|
| BDT1 | 11.3 ±7.0 | 99.7±0.9 | 96.0±7.4 | 4.0±7.4 | 0.0±0.0 |
| BDT2 | **3.4**±0.2 | 100.0±0.0 | **99.5**±1.2 | **0.5**±1.2 | 0.0±0.0 |

Fig. 5 and Fig. 6 depict samples of log likelihood and numbers of DT nodes as well the densities of DT nodes for burn-in and post burn-in phases for the BDT1 and BDT2 strategies. From the top left plot of these figures we can see that the Markov chain very quickly converges to the stationary value of log likelihood near to zero. During post burn-in the values of log likelihood slightly oscillate around zero.

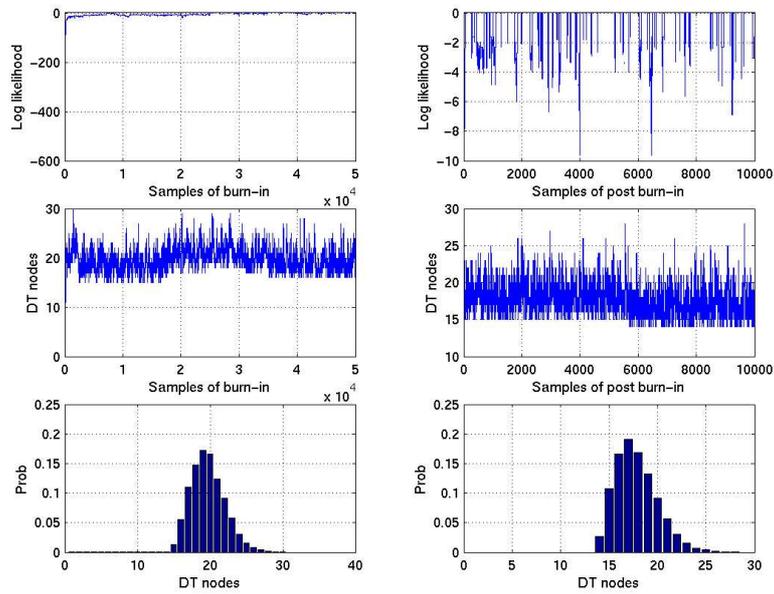

Fig. 5. The Bayesian DT technique with the standard strategy on the XOR3 data: Samples of log likelihood and DT size during burn-in and post burn-in. The bottom plots are the distributions of DT sizes.

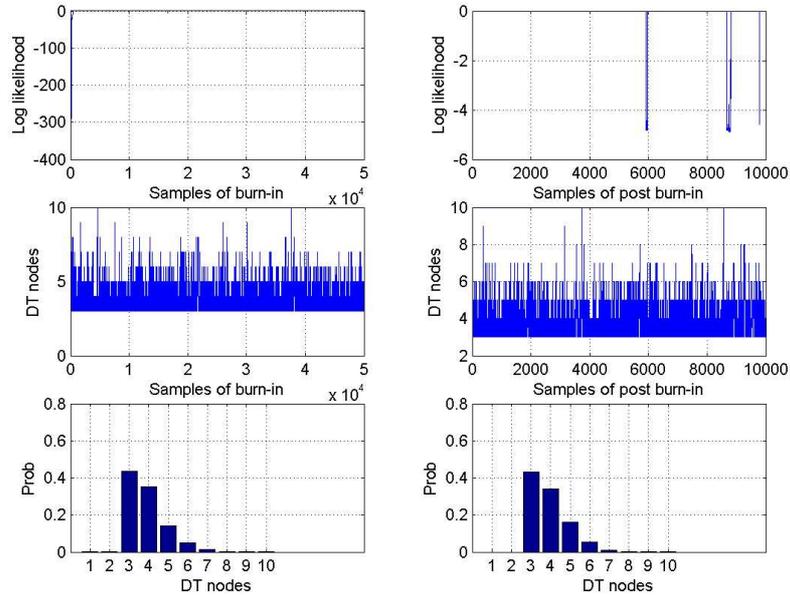

Fig. 6. The Bayesian DT technique with the sweeping strategy on XOR3 problem: Samples of log likelihood and DT size during burn-in and post burn-in. The bottom plots are the distributions of DT sizes.

As we can see from Table 2, both the BDT1 and the BDT2 strategies reveal the same performance on the test data. However the number of DT nodes induced by the BDT2 strategy is much less than that induced by the BDT1 strategy. It is very important that on this test the BDT2 strategy has found a true classification model consisting of the two variables. Besides, the BDT2 strategy provides more sure and correct classifications than those provided by the BDT1 strategy.

### *5.2.2. Performance on Australian Credit Data*

On these data, both the BDT1 and the BDT2 strategies ran with value $p_{min}$ = 3. The numbers of burn-in and post burn-in samples were set equal to 100000 and 10000, respectively. The proposal variance was set equal 1.0.

Both the standard DBT1 and the sweeping BDT2 strategies correctly recognized 85.4% of the test examples. The acceptance rate was 0.5 for the BDT1 and 0.23 for BDT2 strategies. The average number of DT nodes was 25.8 and 8.3 for these strategies, respectively, see Table 3.

Table 3: Comparison between BDT1 and BDT2 on the Australian Credit Data

| Strategy | Number of DT nodes | Perform, % | Sure correct, % | Uncertain, % | Sure incorrect, % |
|---|---|---|---|---|---|
| BDT1 | 25.8 ±2.3 | 85.4±4.0 | 55.1±9.5 | 42.0±9.1 | 2.9±2.9 |
| BDT2 | **8.3**±0.9 | 85.4±4.2 | **65.4**±9.7 | **30.3**±8.9 | 4.3±2.3 |

Table 3 shows us that both the BDT1 and the BDT2 strategies reveal the same performance on the test data. However the number of DT nodes induced by the BDT2 strategy is much less than that induced by the BDT1 strategy. Additionally, the BDT2 strategy provides more sure and correct classifications than those provided by the BDT1 strategy. The rate of uncertain classification is also less than that provided by the BDT1 strategy.

### 5.2.3. Performance on German Credit Data

Both Bayesian strategies ran with value $p_{min}$ = 3. The numbers of burn-in and post burn-in samples were set equal to 100000 and 10000, respectively. The proposal variance was set equal 2.0 to achieve the better performance on these data.

The standard DBT1 and the sweeping BDT2 strategies correctly recognized 72.5% and 74.3% of the test examples, respectively. The acceptance rate was 0.36 for the BDT1 and 0.3 for BDT2 strategies. The average number of DT nodes was 18.5 and 3.8 for these strategies, respectively, see Table 4.

Table 4: Comparison between BDT1 and BDT2 on the German Credit Data

| Strategy | Number of DT nodes | Perform, % | Sure correct, % | Uncertain, % | Sure incorrect, % |
|---|---|---|---|---|---|
| BDT1 | 27.3±2.8 | 72.5±6.8 | 32.8±7.2 | 62.5±11.4 | 4.7±4.4 |
| BDT2 | **20.7**±1.1 | **74.3**±5.9 | **39.4**±9.2 | **54.4**±10.5 | 6.2±3.6 |

As we can see from Table 4, the BDT2 strategy slightly outperforms the BDT1 on the test data. In the same time the number of DT nodes induced by the BDT2 strategy is less than that induced by the BDT1 strategy. The BDT2 strategy provides more sure and correct classifications than those provided by the BDT1 strategy.

### 5.2.4. Performance on Company Liquidity Data

Due to large amount of the training data the BDT1 and the BDT2 strategies ran with value $p_{min}$ = 50. The numbers of burn-in and post burn-in samples were set equal to 50000 and 5000, respectively. The proposal variance was set equal 5.0 which as we found in our experiments provides he best performance.

Both Bayesian DT techniques strategies perform quite well, recognizing 91.5% of the test examples. The acceptance rate was 0.36 for the BDT1 and 0.3 for BDT2 strategies. The average number of DT nodes was 68.5 and 34.2 for these strategies, respectively, see Table 6.

Table 5: Comparison between BDT1 and BDT2 on the Company Liquidity Data

| Strategy | Number of DT nodes | Perform, % | Sure correct, % | Uncertain, % | Sure incorrect, % |
| --- | --- | --- | --- | --- | --- |
| BDT1 | 68.5±5.2 | 91.5±0.3 | 89.8±1.4 | 2.9±2.1 | 7.2±0.8 |
| BDT2 | **34.2**±3.3 | 91.5±0.5 | 90.2±1.1 | 2.5±1.7 | 7.3±0.8 |

Fig. 7 and Fig. 8 depict samples of log likelihood and numbers of DT nodes as well as the densities of DT nodes for burn-in and post burn-in phases for the BDT1 and BDT2 strategies.

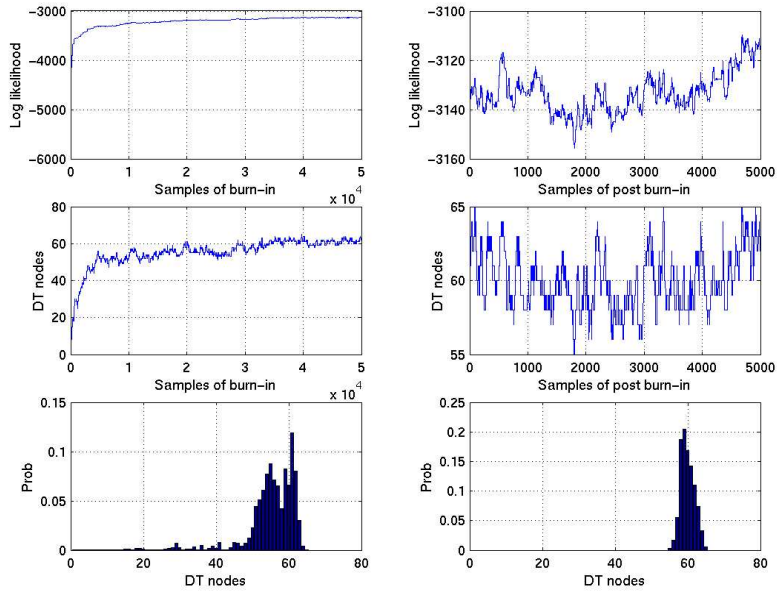

Fig. 7. The Bayesian DT technique with the standard strategy on the Company Liquidity data

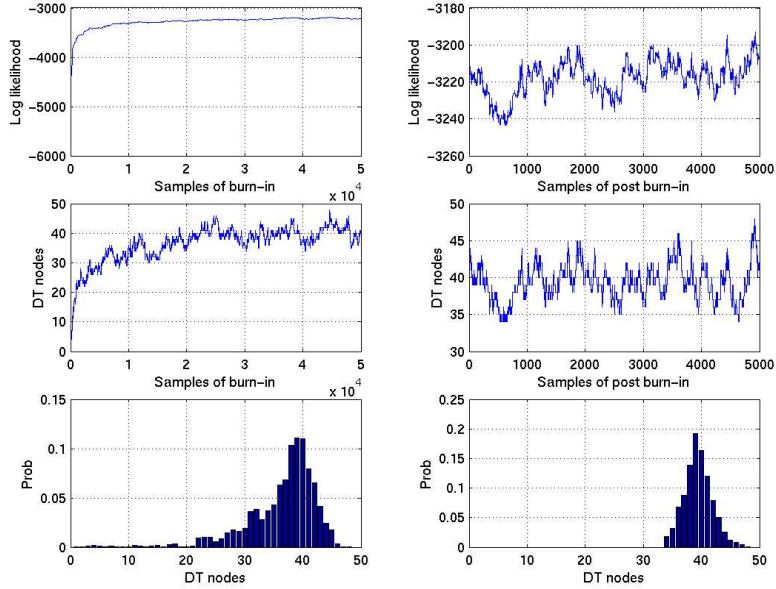

Fig. 8. The Bayesian DT technique with the sweeping strategy on the Company Liquidity.

From Table 5 we can see that both the BDT1 and the BDT2 strategies reveal the same performance on the test data. However the number of DT nodes induced by the BDT2 strategy is much less than that induced by the BDT1 strategy.

## 6.  Conclusion

The use of the RJ MCMC methodology of stochastic sampling from the posterior distribution makes Bayesian DT techniques feasible. However, exploring the space of DTs parameters, existing techniques may prefer sampling DTs from the local maxima of the posterior instead of the properly representing the posterior. This affects the evaluation of the posterior distribution and, as a result, causes an increase in the classification uncertainty. This negative effect can be reduced by averaging the DTs obtained in different starts or by restricting the size of DTs during burn-in phase.

As an alternative way of reducing the classification uncertainty, we have suggested the Bayesian DT technique using the sweeping strategy. Within this strategy, DTs are modified after birth or change moves by removing the splitting nodes containing fewer data points than acceptable.

The performances of the Bayesian DT techniques with the standard and the sweeping strategies have been compared on a synthetic dataset as well as on some datasets from the StatLog Repository and real financial data. Quantitatively evaluating the uncertainty within the Uncertainty Envelope technique, we have found that our Bayesian DT technique using the sweeping strategy is superior to the standard Bayesian DT technique. Both Bayesian DT techniques reveal rather similar average classification accuracy on the test datasets. However, the Bayesian averaging technique with a sweeping strategy makes more sure and incorrect classifications. We also observe that the sweeping strategy provides much shorter DTs.

Thus we conclude that our Bayesian strategy of averaging over DTs using a sweeping strategy is able decreasing the classification uncertainty without affecting classification accuracy on the problems examined. Clearly this is a very desirable property for classifiers used in critical systems in which classification uncertainty may be of crucial importance for risk evaluation.

## Acknowledgments

This research was supported by the EPSRC, grant GR/R24357/01.